\documentclass{llncs}
\usepackage{graphicx}
\usepackage{comment}
\usepackage{amsmath,amssymb} 
\usepackage{color}
\usepackage{gensymb}

\usepackage[ruled,vlined]{algorithm2e}
\usepackage{subfigure}
\usepackage{url}
\usepackage{hyperref}
\usepackage{color}

\usepackage{booktabs}
\usepackage{amsfonts}
\usepackage{times}
\usepackage{multirow}
\usepackage{soul}
\usepackage{tabularx}
\usepackage{array}
\usepackage[misc]{ifsym}
\usepackage[marginal]{footmisc}
\usepackage{bm}
\usepackage{bbm}
\usepackage{amsfonts}

\begin{document}
\pagestyle{headings}
\mainmatter
\def\ECCVSubNumber{1905}  
\def\eg{\emph{e.g.}}
\def\ie{\emph{i.e.}}
\def\etc{\emph{etc}}
\def\wrt{\emph{w.r.t.}}

\title{Bias-based Universal Adversarial Patch Attack\\for Automatic Check-out} 
\titlerunning{ECCV-20 submission ID \ECCVSubNumber}
\author{ Aishan Liu\inst{1}\protect\footnotemark[1] \and
 Jiakai Wang\inst{1}\protect\footnotemark[1] \and
 Xianglong Liu\inst{1,2}\protect\footnotemark[2] \and
 Bowen Cao\inst{1}\and \\
 Chongzhi Zhang\inst{1}\and
Hang Yu\inst{1}} 

\institute{State Key Lab of Software Development Environment, Beihang University, Beijing, China \and
Beijing Advanced Innovation Center for Big Data-Based Precision Medicine,\\ Beihang University, Beijing, China}

\maketitle

\renewcommand{\thefootnote}{\fnsymbol{footnote}} 
\footnotetext[1]{These authors contributed equally to this work.} 
\footnotetext[2]{Corresponding author.} 

\begin{abstract}
Adversarial examples are inputs with imperceptible perturbations that easily misleading deep neural networks (DNNs). Recently, adversarial patch, with noise confined to a small and localized patch, has emerged for its easy feasibility in real-world scenarios.
However, existing strategies failed to generate adversarial patches with strong generalization ability. In other words, the adversarial patches were input-specific and failed to attack images from all classes, especially unseen ones during training. To address the problem, this paper proposes a bias-based framework to generate class-agnostic universal adversarial patches with strong generalization ability, which exploits both the perceptual and semantic bias of models. Regarding the perceptual bias, since DNNs are strongly biased towards textures, we exploit the hard examples which convey strong model uncertainties and extract a textural patch prior from them by adopting the style similarities. The patch prior is more close to decision boundaries and would promote attacks. To further alleviate the heavy dependency on large amounts of data in training universal attacks, we further exploit the semantic bias. As the class-wise preference, prototypes are introduced and pursued by maximizing the multi-class margin to help universal training. Taking Automatic Check-out (ACO) as the typical scenario, extensive experiments including white-box/black-box settings in both digital-world (RPC, the largest ACO related dataset) and physical-world scenario (Taobao and JD, the world’s largest online shopping platforms) are conducted. Experimental results demonstrate that our proposed framework outperforms state-of-the-art adversarial patch attack methods.
\end{abstract}

\section{Introduction}

\begin{figure}[!htb]
	\centering
	\includegraphics[width=0.9\linewidth]{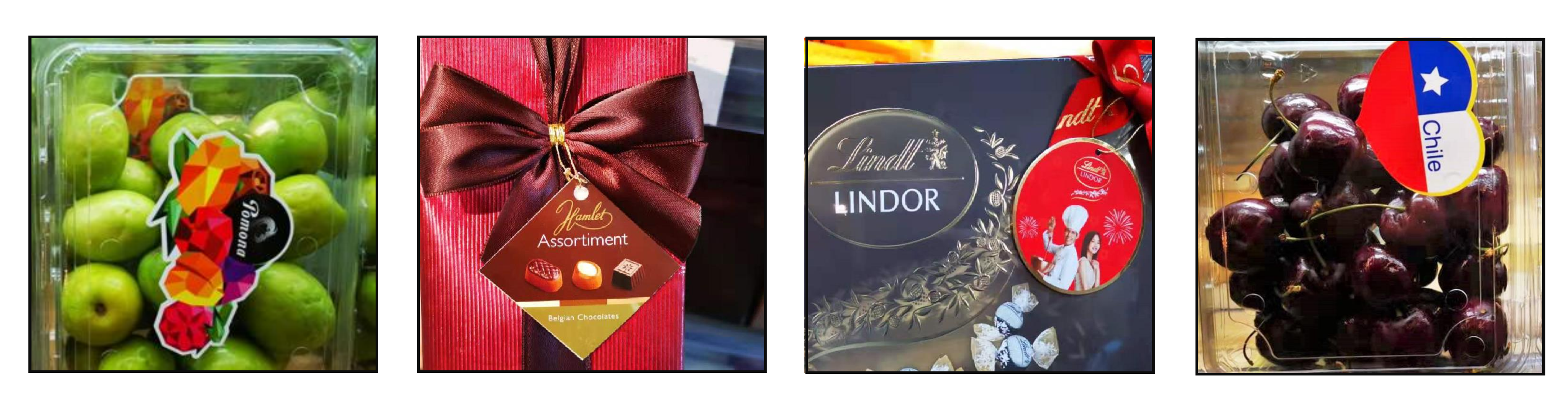}
	\caption{In the real-world scenario like Automatic Check-Out, items (\eg, fruits and chocolates) are often tied with patch-like stickers or tags.}
	\label{fig:firstpage}
\end{figure}

Deep learning has demonstrated remarkable performance in a wide spectrum of areas, including computer vision \cite{krizhevsky2012imagenet}, speech recognition \cite{mohamed2011acoustic} and
natural language processing \cite{sutskever2014sequence}. Recently, deep learning strategies have been introduced into the check-out scenario in supermarkets and grocery stores to revolutionize the way people shopping (\eg, Amazon Go). Automatic Check-Out (ACO) \cite{RPC,li2019data,ekanayake2019naive} is a visual item counting system that takes images of shopping items as input and generates output as a tally of different categories. Customers are not required to put items on the conveyer belt and wait 
for salesclerks to scan them. Instead, they can simply collect the chosen items and a deep learning based visual recognition system will classify them and automatically process the purchase.


Though showing signiﬁcant achievements in our daily lives, unfortunately, deep learning is vulnerable to adversarial examples \cite{goodfellow2014explaining,szegedy2013intriguing}. These small perturbations are imperceptible to human but easily misleading DNNs, which creates potential security threats to practical deep learning applications, \eg, auto-driving and face recognition systems \cite{liu2019perceptual}. In the past years, different types of techniques have been developed to attack deep learning systems \cite{goodfellow2014explaining,szegedy2013intriguing,xiao2018generating}. Though challenging deep learning, adversarial examples are also valuable for understanding the behaviors of DNNs, which could provide insights into the blind-spots and help to build robust models \cite{zhang2019interpreting,zhang2019interpretingand}.

Besides the well-designed perturbations, the adversarial patch serves as an alternative way to generate adversarial examples, which can be directly localized in the input instance \cite{brown2017adversarial,karmon2018lavan,liu2019perceptual}. In contrast, adversarial patches enjoy the advantages of being input-independent and scene-independent. In real-world scenarios, patches could be often observed which are quasi-imperceptible to humans. For example, as shown in Fig.\ref{fig:firstpage}, the tags and brand marks on items in the supermarket. Thus, it is convenient for an adversary to attack a real-world deep learning system by simply generate and stick adversarial patches on the items. However, existing strategies \cite{brown2017adversarial,eykholt2018robust} generate adversarial patches with weak generalization abilities and are not able to perform universal attacks \cite{moosavi2017universal}. In other words, these adversarial patches are input-specific and fail to attack images from all classes, especially unseen ones during training.


To address the problem, this paper proposes a bias-based framework to generate class-agnostic universal adversarial patches with strong generalization ability, which exploits both the perceptual and semantic bias. Regarding the perceptual bias, since DNNs are strongly biased towards textural representations and local patches \cite{zhang2019interpreting,geirhos2018imagenet}, we exploit the hard examples which convey strong model uncertainties and extract a textural patch prior from them by adopting the style similarities. We believe that the textural prior is more close to decision boundaries which would promote the universal attack to different classes.  
To further alleviate the heavy dependency on large amounts of data in training universal attacks\cite{reddy2018ask}, we further exploit the semantic bias. As models have preference and bias towards different features for different classes, prototypes, which contain strong class-wise semantics, are introduced as the class-wise preference and pursued by maximizing the multi-class margin. We generate prototypes to help universal training and reduce the amount of training data required. Extensive experiments including both the white-box and black-box settings in both the digital-world (RPC, the largest ACO related dataset) and physical-world scenario (Taobao and JD, the world’s largest online shopping platforms) are conducted. Experimental results demonstrate that our proposed framework outperforms state-of-the-art adversarial patch attack methods. 

To the best of our knowledge, we are the first to generate class-agnostic universal adversarial patches by exploiting the perceptual and semantic biases of models. With strong generalization ability, our adversarial patches could attack images from unseen classes of the adversarial patch training process or target models. To validate the effectiveness, we choose the automatic check-out scenario and successfully attack the \textbf{Taobao} and \textbf{JD} platform, which are among the world's largest e-commerce platforms and the ACO-like scenarios.

\section{Related work}
\subsection{Adversarial Attacks}
Adversarial examples, which are intentionally designed inputs misleading deep neural networks, have recently attracted research focus \cite{goodfellow2014explaining,szegedy2013intriguing,kurakin2016adversarial}. Szegedy \emph{et al.} \cite{szegedy2013intriguing} first introduced adversarial examples and used the L-BFGS method to generate them. By leveraging the gradients of the target model, Goodfellow \emph{et al.} \cite{goodfellow2014explaining} proposed the Fast Gradient Sign Method (FGSM) which could generate adversarial examples quickly.

To improve the generalization ability to different classes, Moosavi \emph{et al.} \cite{moosavi2017universal} first proposed an algorithm to compute universal adversarial perturbations for DNNs for object recognition tasks. Mopuri \emph{et al.} \cite{mopuri2018generalizable} proposed a data-free objectives to generate universal adversarial perturbations by maximizing the neuron activation. Further, Reddy \emph{et al.} \cite{reddy2018ask} generated data-free universal adversarial perturbations using class impressions.  

Besides, adversarial patch \cite{brown2017adversarial}, with noise confined to a small and localized patch, emerged for its easy accessibility in real-world scenarios. Karmon \emph{et al.} \cite{karmon2018lavan} created adversarial patches using an optimization-based approach with a modified loss function. In contrast to the prior research, they concentrated on investigating the blind-spots of state-of-the-art image classifiers. Eykholt \emph{et al.} \cite{eykholt2018robust} adopted the traditional perturbation techniques to generate attacking noises, which can be mixed into the black and white stickers to attack the recognition of the stop sign. To improve visual fidelity, Liu \emph{et al.} \cite{liu2019perceptual} proposed the PS-GAN framework to generate scrawl-like adversarial patches to fool autonomous-driving systems. Recently, adversarial patches have been used to attack person detection systems and fool automated surveillance cameras \cite{thysvanranst2019}.


\subsection{Automatic Check-out}

The bedrock of an Automatic Check-out system is visual item counting that takes images of shopping items as input and generates output as a tally of different categories \cite{RPC}.
However, different from other computer vision tasks such as object detection and recognition, the training of deep neural networks for visual item counting faces a special challenge of domain shift. Wei \emph{et al.} \cite{RPC} first tried to solve the problem using the data argumentation strategy. To improve the realism of the target images, through a CycleGAN framework \cite{zhu2017unpaired}, images of collections of objects are generated by overlaying images of individual objects randomly. Recently, Li \emph{et al.} \cite{li2019data} developed a data priming network by collaborative learning to determine the reliability of testing data, which could be used to guide the training of the visual item tallying network. 

\section{Proposed Framework}

In this section, we will first give the definition of the problem and then elaborate on our proposed framework.

\subsection{Problem Definition}

Assuming $\mathcal{X}$ $\subseteq \mathbb{R}^{n}$ is the feature space with $n$ the number of features. Supposing ($x_i$ ,$y_i$) is the $i$-th instance in the data with feature vector $x_i$  $\in$  $\mathcal{X}$ and $y_i$ $\in$ $\mathcal{Y}$ the corresponding class label. The deep learning classifier attempts to learn a classification function $F$: $\mathcal{X}$ $\rightarrow$ $\mathcal{Y}$. Specifically, in this paper we consider the visual recognition problem.

An adversarial patch $\delta$ is a localized patch that is trained to fool the target model $F$ to wrong predictions. Given an benign image $x$ with its ground truth label $y$, we form an adversarial example $x'$ which is composed of the original image $x$, an additive adversarial patch $\delta$ $\in \mathbb{R}^z$ and a location mask $M$ $\in$ \{0,1\}$^n$ :
\begin{equation}
x' = (1- M)\odot x + M \odot \delta,
\end{equation}
where $\odot$ is the element-wise multiplication.
The prediction result of $x'_\delta$ by model $F$ is $y'$ = $F$($x'_\delta$). The adversarial patch makes the model predict the incorrect label, namely $y'$ $\neq$ $y$.

To perform universal attacks, we generate a universal adversarial patch $\delta$ that could fool the classifier $F$ on items sampled from distribution $\mu$ from \emph{almost all} classes:
\begin{equation}
F(x) \neq F(x + \delta) \quad \text{for \emph{almost all}} \quad x \sim \mu.
\end{equation}

\subsection{The Framework}

\begin{figure*}[!htb]
	\centering
	\includegraphics[width=1.0\linewidth]{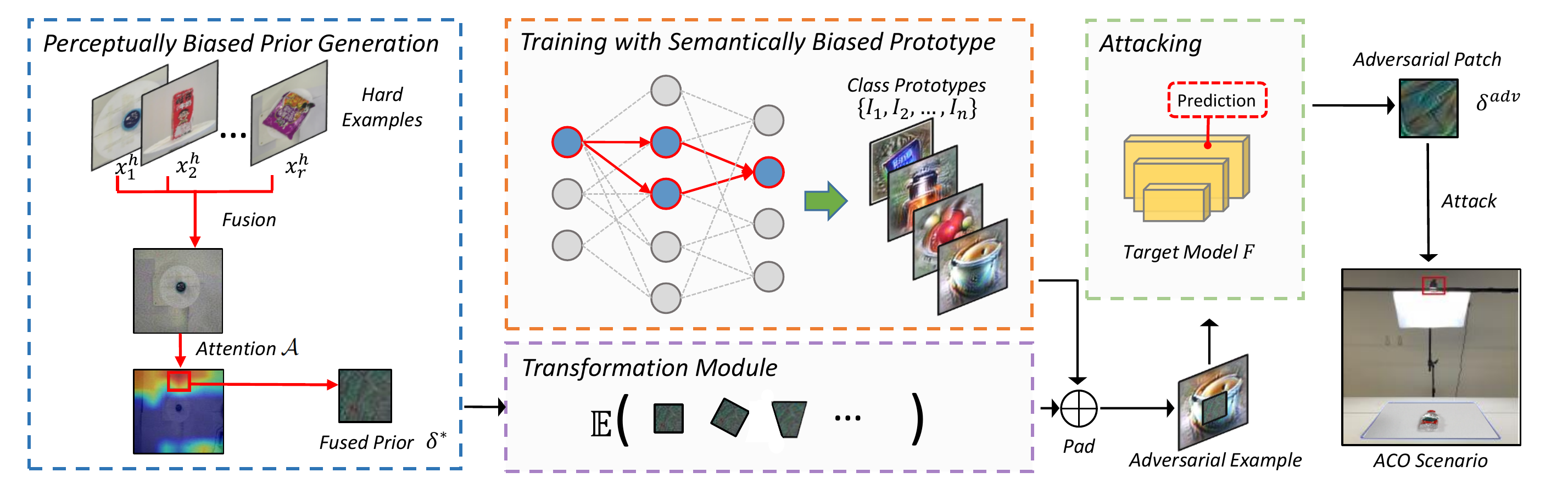}
	\caption{Our bias-based framework to generate universal adversarial patches. We first generate a perceptually biased prior by fusing textural features from multiple hard examples. Then, we generate semantically biased prototypes to help training universal adversarial patches with a target model $F$}
	\label{fig:framework}
\end{figure*}

We propose a bias-based attack framework to generate universal adversarial patches with strong generalization ability. The overall framework can be found in Fig.\ref{fig:framework}.

Recent studies have revealed that DNNs are strongly biased towards texture features when making predictions \cite{geirhos2018imagenet}. 
Deep learning models are still performing well on patch-shuffled images where local object textures are not destroyed drastically \cite{zhang2019interpreting}. 
Thus, we first exploit the perceptual bias of deep models by \textbf{generating perceptually biased priors from multiple hard example set} $\mathcal{X}^h$= $\{x^h_i|i$=$1,...r\}$. Textural features are extracted by an attention module $\mathcal{A}$ to fuse a more powerful prior $\delta^*$. We believe the fused prior are more close to decision boundaries of different classes and would boost universal attacks.


Meanwhile, as models have preferences and impressions for different classes, we further exploit the semantic bias of models for each class. To alleviate the heavy dependency on a large amount of data suffered to train universal attacks, we \textbf{generate semantically biased prototypes to help training}. As the class-wise preference, prototypes contain rich semantics for each class. Thus, prototypes are generated by maximizing the multi-class margin and used to represent instances from each class. Training with prototypes would reduce the amount of training data required. Thus, we generate prototypes \{$I_1$, $I_2$, $...$, $I_n$\} and use them as training data to learn our final adversarial patch $\delta^{adv}$ from $\delta^*$.

\subsection{Perceptually Biased Prior Generation}
Motivated by the fact that deep learning models are strongly biased towards textural features, we first proposed to extract textural features as priors. To fully exploit the statistic uncertainty of models, we borrow textural features from \emph{hard examples}.

Hard examples appear as instances that are difficult for models to classify correctly. Techniques like hard example mining are used to improve training \cite{felzenszwalb2008discriminatively}, in which ``hard'' hence informative examples are spotted and mined. Given a hard example $x^h$ with ground truth label $y$, assuming that $y^h = F(x^h)$ is the prediction of the model $F$. The hard example suffices the constraint that $y^h \ne y$ or with relatively low classification confidence. Obviously, a hard example is an instance lying closely to model decision boundaries, and are more likely to cross the prediction surfaces. Thus, using the features from a hard example $x^h$ to train adversarial patches is like ``standing on the shoulders of a giant'', which would be beneficial to overcome local-optima and gain strong attacking abilities.

To further motivate universal attacks, we extract textural features from multiple hard examples with different labels and fuse them together into a stronger prior. Intuitively, by studying features from multiple hard examples with different labels, our prior would contain more uncertainties for different classes. However, simply learning at pixel-level makes it difficult to extract and fuse textural features. Thus, we introduce the style loss which specifically measures the style differences and encourages the reproduction of texture details:
\begin{equation}
\begin{aligned}
\mathcal{L}_s=\mathbb{E}_k\left[\left\vert\left\vert\mathbf{G}(x^*)-\mathbf{G}(x^h_k)\right\vert\right\vert^2_F\right],\\
\mathbf{G}_{ij}(x)=\sum_k F_{ik}^l(x)\cdot F_{jk}^l(x),
\end{aligned}
\end{equation}

\noindent where $\mathbf{G}$ is the Gram matrix of the features extracted from certain layers of the network. $F_{\cdot k}^l$ is the activation of a specific filter at position $k$ in the layer $l$. $x^*$ is the fused example, and $x^h_k$ is the 
hard example  where $k$= 1, 2, ..., $r$.

Besides, entropy has been widely used to depict the uncertainty of a system or distribution. To further improve universal attacks to different classes, we introduce the class-wise uncertainty loss. we increase model prediction uncertainties by minimizing the negative of entropy. Thus, the fused example would be much closer to decision boundaries and obtain low confidence for different classes. It can be written as: 
\begin{equation}
\begin{aligned}
\mathcal{L}_u=\mathbb{E}_i\left[\log y^{h,i}\right],
\end{aligned}
\end{equation}

\noindent where $y^{h,i}$ denotes the model confidence of the $i$-th class with the fused input $x^*$.

Thus, to fully exploit the perceptual bias, we optimize the fusion loss function $\mathcal{L}_f$ as follows:
\begin{equation}
\begin{aligned}
\mathcal{L}_f=\mathcal{L}_s + \lambda \cdot \mathcal{L}_u,
\end{aligned}
\end{equation}

\noindent where $\lambda$ controls the balance between the two terms.

However, the fused example $x^*$ has a different size with our patches. Thus, an attention module has been introduced to eliminate redundant pixels and generate a textural prior $\delta^*$ from the fused example $x^*$.
\begin{equation}
\begin{aligned}
    & \delta^* = \mathcal{A}(x^*;F),
\end{aligned}
\end{equation}

\noindent where $\mathcal{A}(\cdot)$ is a visual attention module that selects a set of suitable visual pixels from the fused sample. These pixels contain the highest stimuli towards model predictions and would be used as textural priors.

Inspired by \cite{selvaraju2016grad}, given a hard example $x^h$, we compute the gradient of normalized feature maps $Z$ of a specific hidden layer in the model \wrt{} $y^h$. These gradients are global-average-pooled to get the weight matrix which is a weighted combination of feature maps to the hard example $x^h$ :
\begin{equation}
    \begin{aligned}
    a_{ij} = {\sum\limits_{k=1}^w \frac{\partial y^h}{\partial Z_{ij}^k}Z_{ij}^k},
    \end{aligned}
    \label{eqn:aij}
\end{equation}
where $a_{ij}$ represents the weight at position $(i,j)$, $Z_{ij}^k$ is the pixel value in position $(i,j)$ of $k$-th feature map, and $w$ represents the total feature map number. Note that $i\in [0,u-1]$ and $j\in [0,v-1]$ where $u,v$ are the width and height of $Z$, respectively. Then, we can combine the pixels with the highest weight to get our textural prior $\delta^*$.

\subsection{Training with Semantically Biased Prototypes}
With the textural priors generated at the previous stage, we continue to optimize and generate our adversarial patch. To generate universal adversarial perturbations, most of the strategies require a lot of training data \cite{reddy2018ask}. To alleviate the heavy dependency on large amounts of training data, we further exploit the semantic bias.

\emph{Prototype} is a kind of ``quintessential'' observations that best represent and contain the strongest semantics for clusters or classes in a dataset \cite{kim2014bayesian}. Prototypes have provided quantitative benefits to interpret and improve deep learning models. Thus, we further exploit the semantic bias of models (\ie, prototypes)for each class. In this stage, we generate class prototypes and use them during training to effectively reduce the amount of training data required.

Thus, inspired by \cite{simonyan2013deep}, to generate prototypes $I$ representing the semantic preference of a model for each class, we maximize the logits of one specific class. Formally, let $S_t(I)$ denote the logits of class $t$, computed by the classification layer of the target model. By optimizing the \emph{MultiMarginLoss}, the prototype $I_t$ of class $t$ is obtained:

\begin{equation}
\begin{aligned}
I_t=\mathop{\arg\min}_{x} \frac{1}{\mathrm{C}} \sum_{c\neq t} \max(0,margin-S_t(x)+S_c(x))^{p},
\end{aligned}
\label{eqn:classproto}
\end{equation}
where $x$ is the input and satisfies the constraint of an RGB image, $\mathrm{C}$ denotes the total number of classes and $margin$ is a threshold that controls the multi-class margin. In practice, Adam optimizer is applied to find the optimal prototype of class $c$ with $p$ = 1 and $margin$ = 10.

To generate adversarial patches misleading to deep models, we introduce the adversarial attack loss. Specifically, we push the prediction label $y'$ of the adversarial example $x'$ (\ie, a clean prototype $I$ appended with the adversarial patch $\delta^{adv}$) away from its original prediction label $y$. Therefore, adversarial attack loss can be defined as:
\begin{equation}
\begin{aligned}
\mathcal{L}_{t}=  \mathbb{E}_{I,\delta^{adv}}[P(c = t|I') - \max(P(c \neq t|I'))],
\end{aligned}
\label{equ:lt}
\end{equation}
where $\delta^{adv}$ is the adversarial patch which is initialized as the textural prior $\delta^*$, $P(\cdot)$ is the prediction of the target model to the input, $I'$ is the adversarial example which is composed of the prototype $I$ and adversarial patch $\delta^{adv}$, $c$ means the class, and $t$ denotes the class label of $I$.

Moreover, recent studies showed that adversarial examples are ineffective to environmental conditions, \eg, different views, illuminations, \etc. In the ACO scenario, the items are often scanned from different views with different lighting conditions, which would impair the attack ability of our patches. Thus, we further introduce the idea of expectation of transformations to enhance the attack success rate of our adversarial patches in different conditions, as shown in the expectation of conditions $\mathbf{c}$ in Eqn (\ref{equ:lt}).

In conclusion, we first exploit the perceptual bias of models and extract a textural prior from hard examples by adopting the style similarities. To further alleviate the heavy dependency on large amounts of data in training universal attacks, we further exploit the semantic bias. As the class-wise preference, prototypes are introduced and pursued by maximizing the multi-class margin. Using the textural prior as initialization, we train our adversarial patches using the prototypes as training data. The illustration of our two-staged adversarial patch attack algorithm can be found in Algorithm \ref{alg:alg1}.
\begin{algorithm}[t]
\caption{Bias-based Universal Adversarial Patch Attack} \label{alg:alg1}
\KwIn{hard example set $\mathcal{X}^{h}=\{x_i^h|i=1,...,r\}$, target model $F$}
\KwOut{bias-based patch $\delta^{adv}$}
\textbf{Stage1 : Perceptually Biased Prior Generation}\\
initial $x^*$ by randomly select a hard example from $\mathcal{X}^h$\;
\For{the number of fusion epochs}{
    \For{$m = r/batchsize$ steps}{
    sample a minibatch of hard examples from $\mathcal{X}^{h}$\;
    optimize $x^*$ to minimize $\mathcal{L}_f$\;
    }
}
obtain the prior patch $\delta^*$ through attention by Eqn (\ref{eqn:aij})\;
\textbf{Stage2: Training with Semantically Biased Prototype}\\
get class prototypes set $\mathbf{I} = \{I_1,I_2,...I_n\}$ by Eqn (\ref{eqn:classproto})\;
\For{the number of training epochs}{
    \For{$k=n/batchsize $ steps}{
    sample a minibatch of prototypes from $\mathbf{I}$\;
    optimize the adversarial patch $\delta^{adv}$ to minimize $\mathcal{L}_t$ with prototypes\;
    }
}
\end{algorithm}
\section{Experiments}
In this section, we will illustrate the attack effectiveness of our proposed method in different settings in the ACO scenario.

\subsection{Dataset and Evaluation Metrics}
\begin{figure*}
    \centering
    \includegraphics[width=1.0\linewidth]{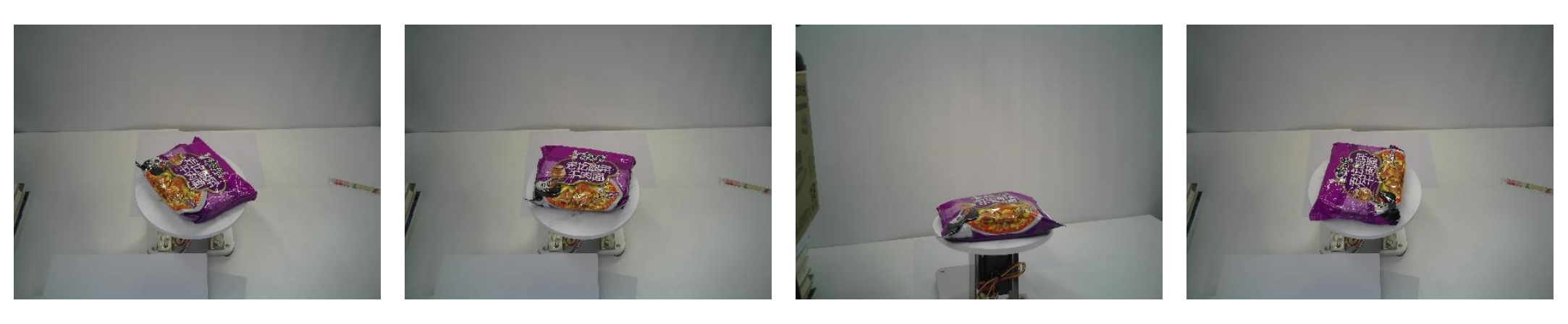}
    \caption{We show 4 images from the category \texttt{instant noodles} in the RPC dataset.}
    \label{fig:database}
\end{figure*}

As for the dataset, we use RPC \cite{RPC}, which is the largest grocery product dataset so far for the retail ACO task. It contains 200 product categories and 83,739 images, including 53,739 single-product exemplary images. Each image is a particular instance of a type of product, which is then divided into 17 sub-categories (\eg, puffed food, instant drink, dessert, gum, milk, personal hygiene, \etc.). Note that the single-product images are taken in an isolated environment and each of them is photographed from four directions to capture multi-views. Fig.\ref{fig:database} shows some images from the RPC dataset. 

To evaluate our proposed method, we choose classification accuracy as the metric. Specifically, we further report $top$-1, $top$-3 and $top$-5 accuracy in our experiments. Note that the goal of adversarial attacks is compromising the performance of the model, \ie, leading to worse values of the evaluation metrics above.

\subsection{Experimental Settings}
The input image is resized to $512 \times 512$ and the patch size is fixed at $32 \times 32$. The size of patches only accounts for 0.38\% of the size of images. To optimize the loss, we use Adam optimizer with a learning rate of 0.01, a weight decay of 10$^{-4}$, and a maximum of 50 epochs. We use 200 hard examples to optimize our fused prior. All of our code is implemented in Pytorch. The training and inference processes are finished on an NVIDIA Tesla k80 GPU cluster.

As for the compared methods, we choose the state-of-art adversarial patch attack methods including AdvPatch \cite{brown2017adversarial}, RP$_2$ \cite{eykholt2017robust}, and PSGAN \cite{liu2019perceptual}. We follow their implementations and parameter settings. Similar to \cite{moosavi2017universal}, we use 50 item samples per class (10,000 in total) as the training data for the compared methods. We also extract 15 prototypes for each class (3,000 in total) as the training data for our method. With respect to the models, we follow \cite{li2019data} for the ACO task and use ResNet-152 as the backbone. 
To further improve the attack success rate of adversarial patches against different environments, we introduce transformations as follows:

- \textbf{Rotation}. The rotation angle is limited in $[-30\degree,30\degree]$.


- \textbf{Distortion}. The distortion rate, \ie, the control argument, moves in $[0, 0.1]$.

- \textbf{Affine Transformation}. The affine rate changes between $0$ and $4$.

\subsection{Digital-world Attack}
In this section, we evaluate the 
performance of our generated adversarial patches on the ACO task in the digital-world in both white-box and black-box settings. We also use a white patch to test the effectiveness of the adversarial attack (denoted as ``White'').

\begin{figure}[!htb]
    \centering
    \subfigure[White-box Attack]{
    \includegraphics[width=0.35\linewidth]{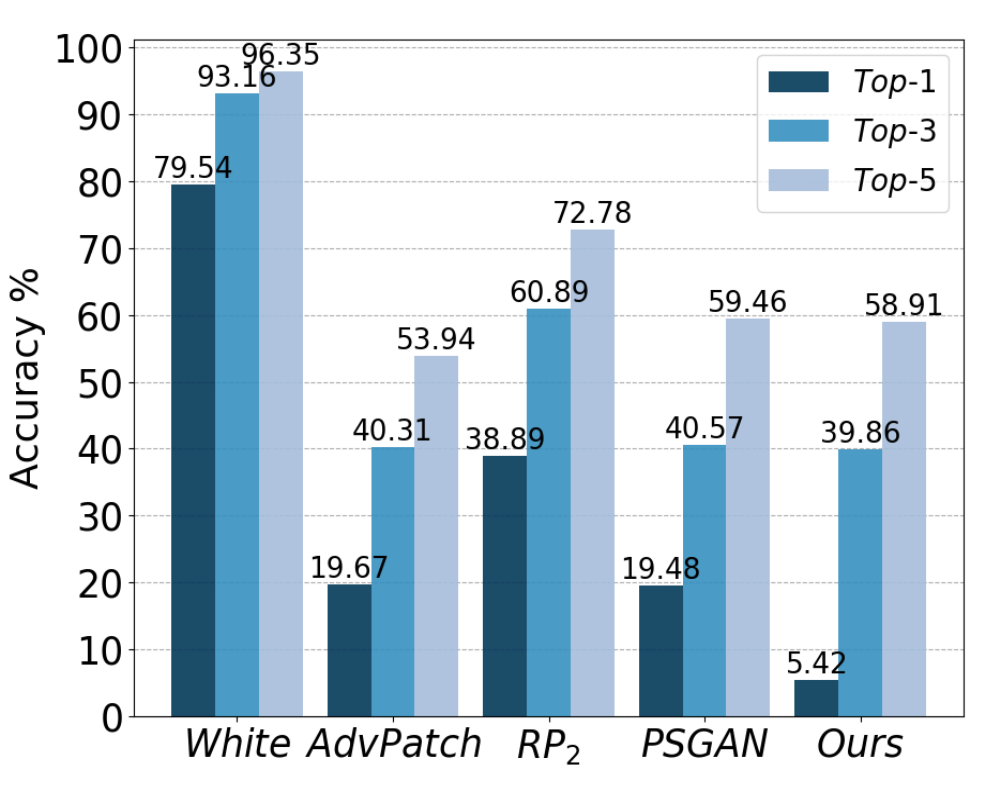}
    \label{fig:whiteboxa}
    }
    \subfigure[Training Process]{
    \includegraphics[width=0.35\linewidth]{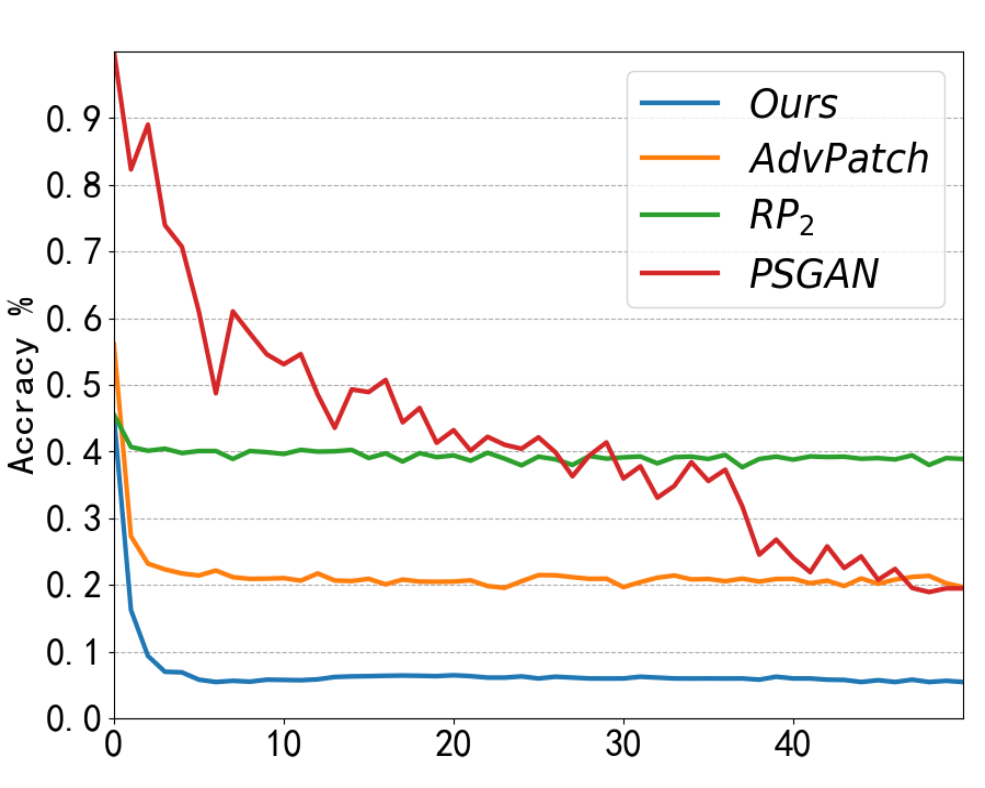}
    \label{fig:whiteboxb}
    }
    \caption{(a) shows the White-box attack experiment in the digital-world with ResNet-152. Our method generates the strongest adversarial patches with the lowest classification accuracy. (b) denotes the training process of different methods}
    \label{fig:whitebox}
\end{figure}

As for the \textbf{white-box} attack, we generate adversarial patches based on a ResNet-152 model and then attack it. As shown in Fig.\ref{fig:whiteboxa}, our method outperforms other compared strategies with large margins. In other words, our adversarial patches obtain stronger attacking abilities with lower classification accuracy contrast to others, i.e., \textbf{5.42\%} to 21.10\%, 19.78\%, and 38.89\% in $top$-1 accuracy. 

As for the \textbf{black-box} attack, we generate adversarial patches based on ResNet-152, then use them to attack other models with different architectures and unknown parameters (\ie, VGG-16, AlexNet, and ResNet-101). As illustrated in Table \ref{tab:blackbox}, our generated adversarial patches enjoy stronger attacking abilities in the black-box setting with lower classification accuracy for different models.

Besides the classification accuracy, Fig.\ref{fig:whiteboxb} shows the training process of adversarial patches using different methods. After several training epochs, the attacking performance of our generated patches becomes stable and keeps the best among all. However, the performance of other methods still vibrates sharply. It is mainly due to the weak generalization ability of other methods. Thus, they achieve different accuracy when attacking different classes showing sharp viberations.

\begin{table}[!b]
    \centering
    \setlength{\tabcolsep}{10pt}
    \caption{Black-box attack experiment in the digital-world with VGG-16, AlexNet, and ResNet-101. Our method generates adversarial patches with strong transferability among different models}
    \begin{tabular}{lllll}
        \toprule
        Model & Method & $top$-1\qquad & $top$-3 & $top$-5 \\ 
        \hline
        \noalign{\smallskip}
        \multirow{4}{*}{VGG-16} & AdvPatch & 73.82 & \textbf{90.73} &\textbf{94.99}  \\
        \cline{2-5} 
        \noalign{\smallskip}
        & RP$_2$ & 81.25& 94.65 & 97.10  \\
        \cline{2-5} 
        & PSGAN & 74.69 & 91.25 & 96.12 \\
        \cline{2-5} 
        \noalign{\smallskip}
        & \textbf{Ours} & \textbf{73.72} & 91.53 & 95.43  \\
        \hline
        \noalign{\smallskip}
        \multirow{4}{*}{AlexNet} & AdvPatch & 51.11 & 72.37 &79.79  \\
        \cline{2-5} 
        \noalign{\smallskip}
        & RP$_2$ & 68.27 & 86.49 & 91.08  \\
        \cline{2-5} 
        \noalign{\smallskip}
        & PSGAN & 49.39 & 72.85 & 82.94  \\
        \cline{2-5} 
        \noalign{\smallskip}
        & \textbf{Ours} & \textbf{31.68} & \textbf{50.92} & \textbf{60.19}  \\
        \hline
        \noalign{\smallskip}
        \multirow{4}{*}{ResNet-101} & AdvPatch & 56.19 & 80.99 &91.52  \\
        \cline{2-5} 
        \noalign{\smallskip}
        & RP$_2$ & 73.52 & 93.75 & 98.13  \\
        \cline{2-5} 
        \noalign{\smallskip}
        & PSGAN & 51.26 & 79.22 & 90.47 \\
        \cline{2-5} 
        \noalign{\smallskip}
        & \textbf{Ours} & \textbf{22.24} & \textbf{51.32} & \textbf{60.28}  \\
        \bottomrule
    \end{tabular}
    
    \label{tab:blackbox}
\end{table}

\subsection{Real-world Attack}
\begin{figure}[!tb]
    \centering
    \subfigure[Taobao]{
    \includegraphics[width=0.35\linewidth]{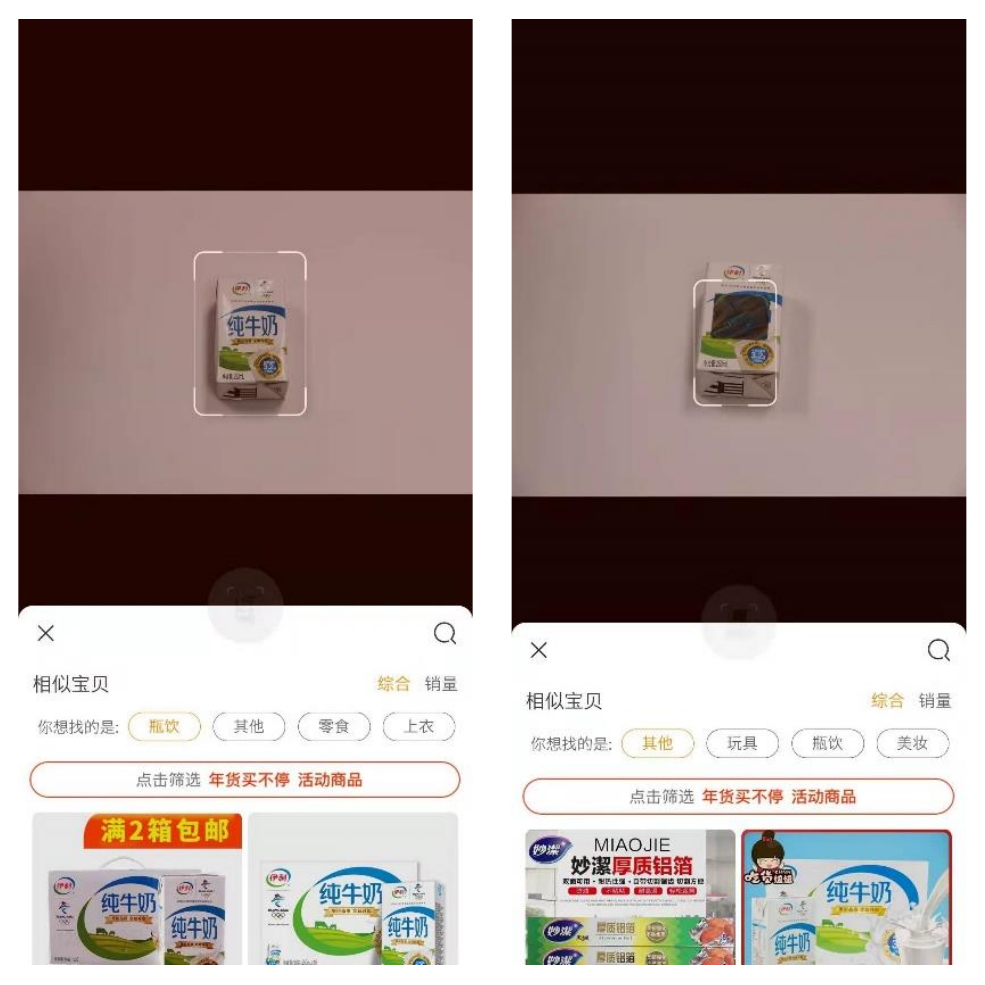}
    }
    \subfigure[JD]{
    \includegraphics[width=0.35\linewidth]{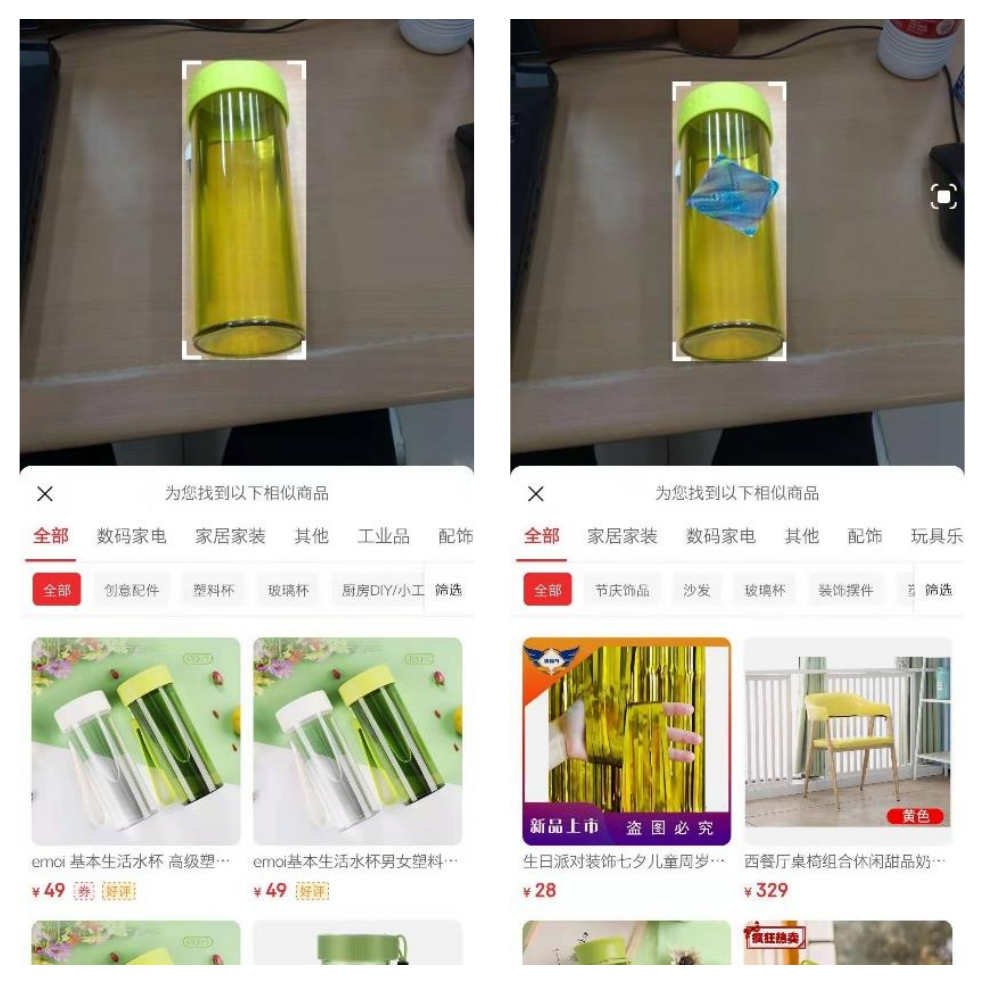}
    }
    \caption{Attack Taobao and JD platform with our adversarial patches. The \texttt{milk} in (a) and the \texttt{plastic cup} in (b) are recognised as the \texttt{decorations} and the \texttt{aluminum foil} when we stick our adversarial patches, respectively}
    \label{fig:real-world}
\end{figure}

To further validate the practical effectiveness of our generated adversarial patches, a real-world attack experiment is conducted on several online shopping platforms to simulate the ACO scenario. We use \textbf{Taobao} (\href{https://market.m.taobao.com/}) and {\textbf{JD}} (\href{https://app.jd.com/}), which are among the biggest e-commerce platforms in the world. We take 80 pictures of 4 different real-world products with different environmental conditions (\ie, angles \{-30\degree, -15\degree, 0\degree, 15\degree, 30\degree\} and distances \{0.3m, 0.5m, 0.7m, 1m\}). 
The $top$-1 classification accuracy of these images is 100\% on Taobao and 95.00\% on JD, respectively. Then, we print our adversarial patches by an HP Color LaserJet Pro MFP M281fdw printer, stick them on the products and take photos with the combination of different distances and angles using a Canon D60 camera. 
A significant drop in accuracy on both platforms can be witnessed with low classification accuracy (\ie, \textbf{56.25\%} on Taobao, \textbf{55.00\%} on JD). The results demonstrate the strong attacking ability of our adversarial patches in real-world scenarios on practical applications. Visualization results can be found in Fig.\ref{fig:real-world}.

\subsection{Generalization Ability}
In this section, we further evaluate the generalization ability of adversarial patches on unseen classes. We perform two experiments using the backbone model (ResNet-152), including attacking unseen item classes of adversarial patch training process and target models. For attacking unseen classes of the patch training process, we first train patches on a subset of the dataset, \ie, only images from 100 classes are used \wrt{} 200 classes (we use prototypes for our method and item images for compared methods). According to the results in Table \ref{tab:us}, our framework generates adversarial patches with strong generalization ability and outperforms other compared methods with huge margins (\ie, \textbf{7.23\%} to 40.28\%, 31.62\%, and 60.87\%). Meanwhile, we also tested the generalization ability on classes that have never been ``seen'' by the target model. Specifically, we train our patches on the RPC dataset and test them on the Taobao platform. We select 4 items and stick adversarial patches on them and take 64 pictures. The categories of the items are unseen to target models (not in the 200 classes for ResNet-152), but known to the Taobao platform.
Interestingly, after attacks, the $top$-$1$ accuracy on Taobao is $\textbf{65.63\%}$. Though our patches are not trained to attack some specified classes, they still generalize well to these unseen classes. Thus, we can draw the conclusion that our framework generates universal adversarial patches with strong generalization abilities to even unseen classes. 
\begin{table}[!htb]
\centering
\begin{center}
\caption{Attack on unseen classes. Our method generates adversarial patches with the strongest generalization abilities showing lowest accuracy compared with other methods}
\setlength{\tabcolsep}{15pt}{
\begin{tabular}{lllll}
\toprule
Method & AdvPatch & RP$_2$ & PSGAN & \textbf{Ours} \\
\midrule
$top$-1 & 40.28 & 60.87 & 31.62 & \textbf{7.23} \\
\bottomrule
\end{tabular}}
\label{tab:us}
\end{center}
\end{table}

\subsection{Analysis of Textural Priors}
Since textural priors have improved universal attacks, a question emerges: ``Why and how the textural priors are beneficial to universal adversarial attacks?'' Thus, in this section, we further study the effectiveness of our textural priors.

\subsubsection{Training from Different Priors}  
\begin{figure}[!b]

	\subfigure[Different Priors]{
	\includegraphics[width=0.38\linewidth]{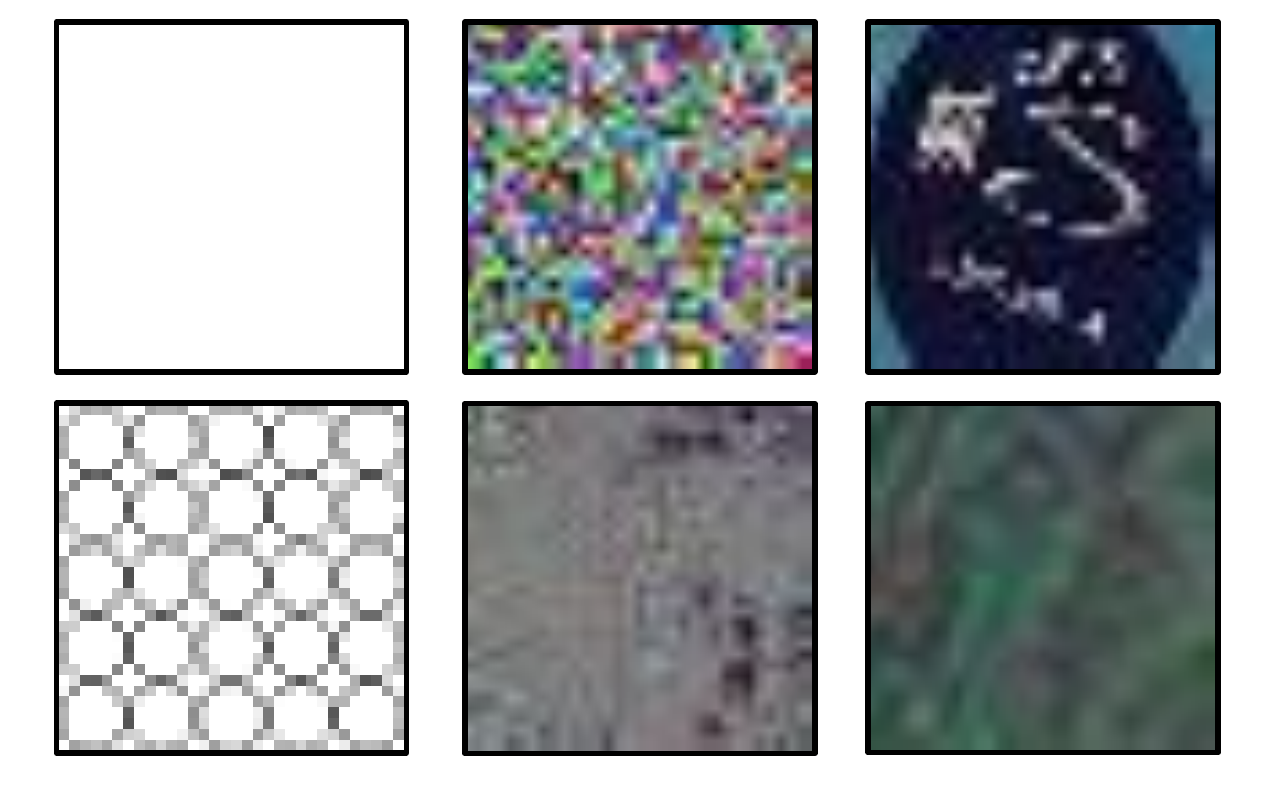}
	\label{fig:diffpriors}
	}
	\hspace{5mm}
	\subfigure[Boundary Distance]{
	\includegraphics[width=0.32\linewidth]{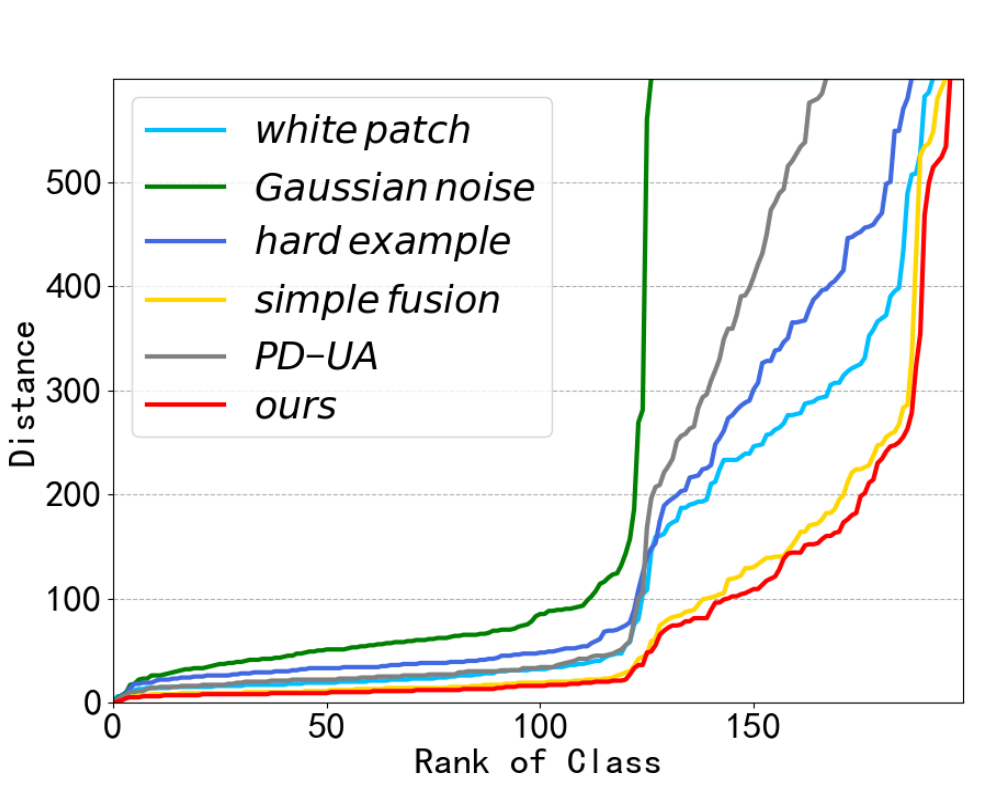}
	\label{fig:db}
	}
    \centering
	\caption{(a) shows different priors we used to generate adversarial patches. They are white patch, Gaussian noise, hard example, PD-UA, simple fusion, and our \textbf{textural prior} respectively, from up to down, left to right. (b) is the decision boundary distance analysis, where our fused prior achieves the smallest decision boundary distances for each class}
	\label{fig:initialization}
\end{figure}

To demonstrate the effectiveness of our textural priors, we begin to study by initializing patches through different priors, \eg, white patch, Gaussian noise, hard example, PD-UA \cite{liu2019universal}, simple fusion, and our textural prior (denoted as ``ours''). In contrast to our textual prior, we use the same amount of simple examples to generate the simple version of fused prior (denoted as ``SimpleFusion''). Other experimental settings are the same as the settings of the digital-world attack. The visualization of them can be found in Fig.\ref{fig:diffpriors}. We train 6 adversarial patches with all the same experimental settings except for the initialization. The corresponding accuracy after attacking are 17.67\%, 18.96\%, 16.11\%, 21,10\%, 24.09\%, \textbf{5.42\%}. It shows that our fused priors offer adversarial patches the strongest attacking ability. 


\subsubsection{Decision Boundary Distance Analysis} 
The minimum distance to the decision boundary among the data points reflects the model robustness to small noises \cite{cortes1995support}. Similarly, the distance to decision boundaries for an instance characterizes the feasibility performing attack from it.
Due to the computation difficulty of decision boundary distance for deep models, we calculate the distance of an instance $x$ to specified classes \wrt{} the model prediction to represent the decision boundary distance. Given a learnt model $F$ and point $x_i$ with class label $y_i$ ($i=1,\ldots,N$), for each direction ($y_j$,$i \neq j$) we estimate the smallest step numbers moved as the distance. We use the $\mathrm{L}_2$ norm Projected Gradient Descent (PGD) until the model's prediction changes, i.e., $F(x_i) \neq y_i$.

As shown in Fig.\ref{fig:db}, our textural priors obtain the minimum distance in each direction compared to other initialization strategies. It explains the reason that our textural prior performs stronger adversarial attacks because it is more close to the decision boundaries.

\subsection{Ablation Study}
In this section, we investigate our method through ablation study. 




\subsubsection{The Effectiveness of Class Prototypes}
In this section, we evaluate the effectiveness of the prototypes using ResNet-152. We first study it by using different amounts of prototypes. Specifically, we mix class prototypes and item images from the RPC dataset in different ratios. We use them to train different adversarial patches and assess their attacking ability. Note that the total number of the training data is fixed as 1000. As shown in Table \ref{tab:proto}, with the increasing number of prototypes, adversarial patches becoming stronger (\ie, lower accuracy). 

Besides, we further investigate the amount of data required with our framework to train adversarial patches by solely using prototypes or item images, respectively. Specifically, we first train adversarial patches with 1000 prototypes as Ours$_{P1000}$. Then, we randomly select 1000, 2000, 4000, 10000 item images from the RPC dataset to train adversarial patches, respectively (denoted by Ours$_{I1000}$, Ours$_{I2000}$, Ours$_{I4000}$, and Ours$_{I10000}$). The results in Table \ref{tab:proto2} show that to achieve the approximate attacking ability in Ours$_{P1000}$ setting, a lot more items images are required. It indicates the representative ability of class prototypes for different classes. Thus, we can conclude that our class prototypes are beneficial to improve the attack ability and reduce the amount of training data required.

\begin{table}[!htb]
    \centering
    \setlength{\tabcolsep}{10pt}
    \caption{The $top$-1 accuracy of the adversarial patches obtained using different amount of training data. Our prototypes need only half the data to achieve a similar performance compared to original method.}
    \begin{tabular}{llllll}
    \toprule
    Settings & Ours$_{P1000}$ & Ours$_{I1000}$ & Ours$_{I2000}$ & Ours$_{I4000}$ & Ours$_{I10000}$ \\
    \hline
    \noalign{\smallskip}
    $top$-$1$ & \textbf{6.51} & 12.43 & 6.57 & 6.10 & 5.40 \\
    \bottomrule
    \end{tabular}
    
    \label{tab:proto2}
\end{table}

\begin{table}[h!]
\begin{minipage}[b]{0.45\linewidth}
    \centering
    \setlength{\tabcolsep}{5pt}
    \caption{Training with a different mixture of class prototypes and original item images using the same amount of training data. Obviously, training with class prototypes give adversarial patches with the strongest attack ability.}
    \begin{tabular}{ll}
        \toprule
        Mixture settings & $top$-$1$  \\
        (\#Prototypes : \#Item images)&  \\
        \hline
        \noalign{\smallskip}
        1000 : 0 & \textbf{6.51} \\
        \hline
        \noalign{\smallskip}
        750 : 250 & 7.81 \\
        \hline
        \noalign{\smallskip}
        500 : 500 & 10.03 \\
        \hline
        \noalign{\smallskip}
        250 : 750 & 11.55 \\
        \hline
        \noalign{\smallskip}
        0 : 1000 & 12.43 \\
        \bottomrule
    \end{tabular}
    \label{tab:proto}
\end{minipage}
\hspace{10mm}
\begin{minipage}[b]{0.45\linewidth}
    \centering
    \setlength{\tabcolsep}{5pt}
    \caption{Ablation study on transformation module. Note that setting \textbf{w/} means the patch is generated with transformation in training progress, and the setting \textbf{w/o} is the opposite. All of the generated patches are tested in the digital world.}
    \begin{tabular}{lll}
        \toprule
        Transformation & Settings & $top$-$1$ \\
        \hline
        \noalign{\smallskip}
        \multirow{2}{*}{Rotation} & w/ & \textbf{11.98}  \\
        \cline{2-3}
        \noalign{\smallskip}
        & w/o & 13.01\\
        \hline
        \noalign{\smallskip}
        \multirow{2}{*}{Distortion} & w/ & \textbf{22.26} \\
        \cline{2-3}
        \noalign{\smallskip}
        & w/o & 30.70\\
        \hline
        \noalign{\smallskip}
        \multirow{2}{*}{Affine} & w/ & \textbf{16.38}  \\
        \cline{2-3}
        \noalign{\smallskip}
        & w/o & 21.10\\
        \bottomrule
    \end{tabular}
    \label{tab:transform}
\end{minipage}
\end{table}

\subsubsection{Transformation Module}
Studies have shown that adversarial examples are ineffective to environmental conditions, \eg, different rotations, illuminations, \etc. In the ACO scenario, items are often scanned from different views with different lighting conditions. Thus, we introduce a transformation module to reduce the impact of environmental conditions to the the attack ability. Here, we study the effectiveness of different transformation types we used in the module. Specifically, we employ ResNet-152 as the target model and execute only one kind of transformation in each experiment. As shown in Table \ref{tab:transform}, enabling transformations would increase attacking ability in ACO scenario with lower accuracy (\ie,\textbf{11.98\%} to 13.01\% in rotation setting, \textbf{22.26\%} to 30.70\%  in distortion setting, \textbf{16.38\%} to 21.10\% in affine setting).

    

\section{Conclusions}

In this paper, we proposed a bias-based attack framework to generate class-agnostic universal adversarial patches, which exploits both the perceptual and semantic bias of models. Regarding the perceptual bias, since DNNs are strongly biased towards textures, we exploit the hard examples which convey strong model uncertainties and extract a textural patch prior from them by adopting the style similarities. The patch prior is more close to decision boundaries and would promote attacks. To further alleviate the heavy dependency on large amounts of data in training universal attacks, we further exploit the semantic bias. As the class-wise preference, prototypes are introduced and pursued by maximizing the multi-class margin to help universal training.  Taking  ACO as the typical scenario, extensive experiments are conducted which demonstrate that our proposed framework outperforms state-of-the-art adversarial patch attack methods. 

Since our adversarial patches could attack the ACO system, it is meaningful for us to study how and why DNNs make wrong predictions. Our framework provides an effective path to investigate model blind-spots. Beyond, it could also be beneficial to improve the robustness of ACO systems in practice.

Model biases, especially texture-based features, have been used to perform adversarial attacks. In contrast, we are also interested in improving model robustness from the perspective of model bias. Can we improve model robustness by eliminating the textural features from the training data? We leave it as future work.

\section{Acknowledge}

This work was supported by National Natural Science Foundation of China (61872021, 61690202)，Beijing Nova Program of Science and Technology (Z191100001119050), and Fundamental Research Funds for Central Universities (YWF-20-BJ-J-646).
\clearpage
%
%
\bibliographystyle{splncs04}
\bibliography{egbib}
\end{document}